%% file: main.tex
\documentclass{ecai} 

\usepackage{latexsym}
\usepackage{amssymb}
\usepackage{amsmath}
\usepackage{amsthm}
\usepackage{booktabs}
\usepackage{enumitem}
\usepackage{graphicx}
\usepackage{color}

\usepackage[table,xcdraw]{xcolor}
\usepackage{multirow}
\usepackage[normalem]{ulem}
\useunder{\uline}{\ul}{}

\newcommand{\BibTeX}{B\kern-.05em{\sc i\kern-.025em b}\kern-.08em\TeX}
\usepackage{subcaption}

\begin{document}


\begin{frontmatter}


\paperid{102} 



\title{FlowLearn: Evaluating Large Vision-Language Models on Flowchart Understanding}


\author[A]{\fnms{Huitong}~\snm{Pan}\thanks{Corresponding Author. Email: huitong.pan@temple.edu}}
\author[A]{\fnms{Qi}~\snm{Zhang}}
\author[B]{\fnms{Cornelia}~\snm{Caragea}}
\author[A]{\fnms{Eduard}~\snm{Dragut}}
\author[A]{\fnms{Longin}~\snm{Jan Latecki}}

\address[A]{Dept. of Computer and Information Sciences, Temple University, Philadelphia}
\address[B]{Dept. of Computer Science, University of Illinois Chicago}


\begin{abstract}
Flowcharts are graphical tools for representing complex concepts in concise visual representations. This paper introduces the FlowLearn dataset, a resource tailored to enhance the understanding of flowcharts. FlowLearn contains complex scientific flowcharts and simulated flowcharts. The scientific subset contains 3,858 flowcharts sourced from scientific literature and the simulated subset contains 10,000 flowcharts created using a customizable script. The dataset is enriched with annotations for visual components, OCR, Mermaid code representation, and VQA question-answer pairs. Despite the proven capabilities of Large Vision-Language Models (LVLMs) in various visual understanding tasks, their effectiveness in decoding flowcharts—a crucial element of scientific communication—has yet to be thoroughly investigated. The FlowLearn test set is crafted to assess the performance of LVLMs in flowchart comprehension. Our study thoroughly evaluates state-of-the-art LVLMs, identifying existing limitations and establishing a foundation for future enhancements in this relatively underexplored domain. For instance, in tasks involving simulated flowcharts, GPT-4V achieved the highest accuracy (58\%) in counting the number of nodes, while Claude recorded the highest accuracy (83\%) in OCR tasks. Notably, no single model excels in all tasks within the FlowLearn framework, highlighting significant opportunities for further development. 

\end{abstract}
    
\end{frontmatter}


\section{Introduction}
\begin{figure*}
    \centering
\includegraphics[width=\textwidth]{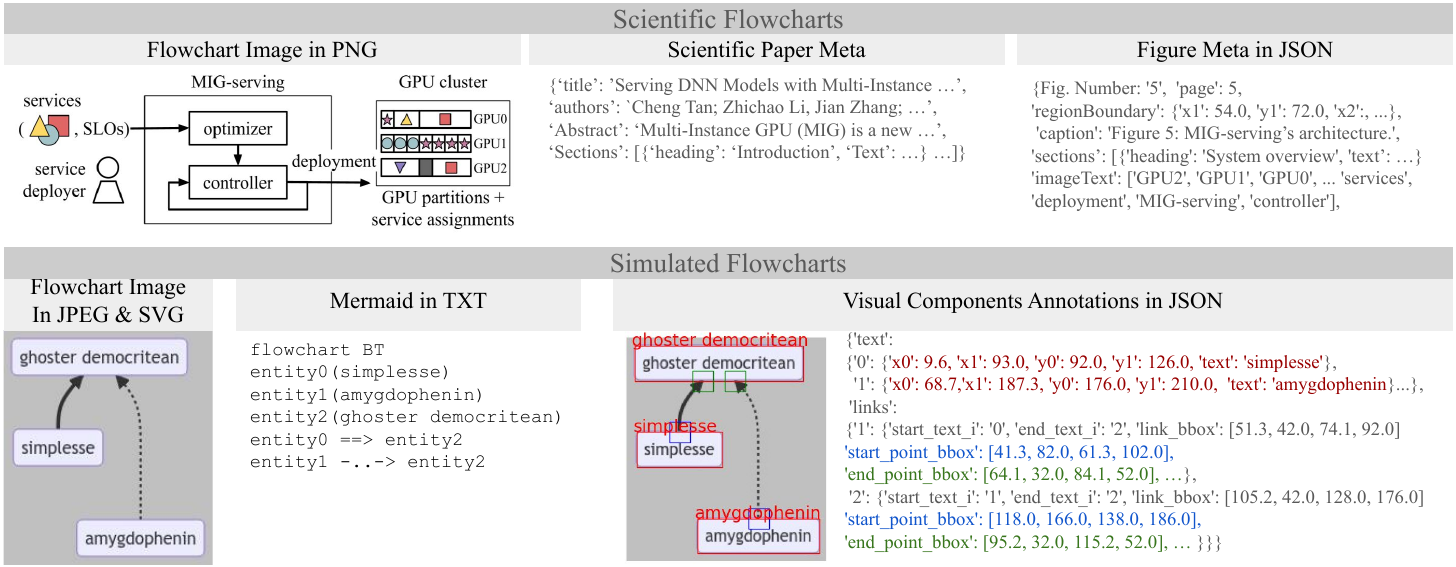}
\caption{Overview of the FlowLearn Dataset illustrating the detailed components within the Scientific and Simulated subsets.}
\label{fig:meta}
\end{figure*}

Flowcharts are visual tools that simplify complex processes and concepts across various domains, condensing intricate information into concise visual representations that enhance both comprehension and communication. In this paper, a flowchart is defined as a diagram that outlines a sequence of operations using standardized symbols like rectangles for steps and arrows to indicate process flow, as demonstrated in Figure \ref{fig:meta}.

Flowchart comprehension, particularly in the domain of computer vision and LVLMs, remains underexplored despite their extensive application. Resources like ACL-Fig~\cite{aclfig} that include scientific flowcharts are limited and often provide only basic figure captions and sparse inline reference annotations. Flowcharts' complex nature—requiring text recognition, identification of various visual elements (e.g., boxes, nodes, symbols), and understanding of node connections—demands more comprehensive annotations for effective evaluation, emphasizing the need for specialized resources.

Further underscoring the inadequacy of current resources, our preliminary analysis of 208 flowcharts from ACL-Fig using Gemini-Pro-Vision~\cite{gemini} yielded a low BLEU score of 0.006 (refer to supplementary material for details). However, this score doesn't fully represent the model’s comprehension capabilities. The captions in ACL-Fig, sometimes as brief as `Figure 2: Alignment learning algorithm', do not provide robust ground truths for meaningful evaluation. With a median caption length of just nine words, ACL-Fig is inadequate for evaluating flowchart understanding. 

Addressing these gaps, we introduce the FlowLearn Dataset\footnote{Our dataset is available on \url{https://huggingface.co/datasets/jopan/FlowLearn}. Our code for generation and evaluation is available on \url{https://github.com/Jo-Pan/FlowLearn}}, which includes both scientific and simulated flowcharts. The scientific subset features 3,858 flowcharts sourced from scientific literature, annotated with captions (median length of 25 words) and in-figure text. The simulated subset consists of 10,000 flowcharts generated from Mermaid code. 
This simulated subset enhances the dataset by providing detailed annotations of visual components, thereby enabling quantitative evaluations of component-specific tasks. Additionally, both subsets include Visual Question Answering (VQA) question-answer pairs, further enhancing their utility for training and model evaluation.

In addition to introducing a novel dataset tailored for enhancing flowchart comprehension, this paper provides a rigorous analysis of the performance of contemporary LVLMs in interpreting flowcharts. Our findings reveal significant room for improvement in LVLMs, with no single model excelling across all tasks within the FlowLearn framework. For instance, in tasks involving simulated flowcharts, GPT-4V achieved the highest accuracy (58\%) in counting the number of nodes, while Claude recorded the highest accuracy (83\%) in OCR tasks. This varied performance highlights specific areas where LVLM capabilities could be further developed. Given the rapid advancements in the fields of Large Language Models (LLMs) and LVLMs, FlowLearn is both timely and valuable, providing a foundation for future research in visual data interpretation and automated reasoning, and setting new benchmarks in the field.

\section{Related Works}
This section overviews interdisciplinary research at the intersection of computer vision and natural language processing, focusing on the comprehension of visual figures in scientific contexts.

\vspace{-10pt}
\subsection{Scientific Figure Datasets}
Significant efforts have been made to develop methodologies and datasets aimed at extracting and understanding scientific figures. Methods such as PDFigCapX \cite{pdffigcapx}, PDFMEF \cite{pdfmef}, PDFFIGURES \cite{pdffigures}, 
and PDFFIGURES2 \cite{pdffigures2} facilitate the extraction of figures, captions, and related information from scholarly articles. Whereas datasets like VIS30K \cite{vis30k} and PDFFIGURES2 
advance figure extraction by providing detailed annotations for figure locations. Moreover, ACL-Fig \cite{aclfig} and DocFigure \cite{docfigure} focus on figure classification, enhancing the understanding of various figure types, including bar charts and architecture diagrams. Additionally, specialized datasets like SciCap \cite{scicap} and Parsing-AUC \cite{parsing_auc} concentrate on image captioning and summarization for experimental results figures.

Despite advancements, a significant gap exists in datasets with detailed annotations for flowcharts. For instance, ACL-Fig contains only about 208 flowcharts, primarily as architecture diagrams and neural networks, often duplicated, with brief captions or blurry figures from pre-2000 papers. CSDia \cite{csdia} focuses on logical diagrams but lacks detailed caption information as it is not derived from the scientific literature.

\subsection{Visual Understanding Resources and Methods}
Research in image captioning~\cite{review_visual2text, review_visual2text2}, particularly in scientific chart image captioning~\cite{review_auto_chart}, has seen significant advancements, exemplified by works such as Parsing-AUC \cite{parsing_auc}, which combines figure semantics extracted via OpenCV with textual information from the main text to generate comprehensive figure summaries for AUC figures.

The field of VQA \cite{review_vqa, review_image_captioning} 
has progressed substantially, with datasets like VL-ICL Bench \cite{vlicl} providing benchmarks for multimodal in-context learning. In scientific contexts, CSDia \cite{csdia} employs models based on Diagram Parsing Nets to tackle VQA tasks. 
In scientific contexts, FigureSeer \cite{figureseer} automates figure localization, classification, and summarization. It extracts key visual components such as axes, legends, and data points for analysis, emphasizing the importance of figure decomposition in enhancing the understanding of scientific figures.

In non-scientific contexts, innovative approaches such as Neural-Symbolic VQA \cite{neural_symbolic_vqa} and $\alpha$ILP \cite{alphailp} have been developed to transform neural network outputs into symbolic representations, which are then used to formulate answers. Additionally, the research highlighted in \cite{structure_aware_viz} stresses the importance of incorporating structural information in visualization retrieval processes. Their findings indicate a marked preference among survey participants for evaluating similarity based on visual elements rather than merely pixel-level details, suggesting a deeper, more structural approach to image analysis that could greatly benefit VQA systems.

Several works have focused on extracting objects and their relationships from scientific figures. Notably, \cite{image_to_markup} pioneers the conversion of scientific equation images into LaTeX format. ChartDetective \cite{chartdetective} introduces an interactive application for converting result chart images into SVG, preserving semantics and component relationships. However, it is essential to note that this approach relies on user interaction and takes about 4 minutes for a single conversion. For non-scientific domains, Flow2Code \cite{Flow2Code} converts hand-drawn flowcharts to simple code by using object detection and rules.

The development of LVLMs has marked a significant leap in visual understanding, with models capable of integrating advanced vision techniques with LLMs. These models can learn simultaneously from images and texts, tackling various tasks such as visual question answering and image captioning. The OpenCompass Multi-modal Leaderboard\footnote{https://rank.opencompass.org.cn/leaderboard-multimodal} rank these LVLMs, includes entries such as GPT-4V~\cite{gpt4v}, Gemini~\cite{gemini}, LLaVA~\cite{llava}, Claude~\cite{claude},  InternLM~\cite{internlm}, Qwen-VL~\cite{qwenvl}, Step-1V\footnote{https://www.stepfun.com/}, and DeepSeek~\cite{deepseek}.
All of these models were trained on diverse datasets, including datasets for VQA, optical character recognition (OCR), and academic-related VQA. These are essential building blocks for achieving flowchart comprehension and enhancing the understanding of scientific flowcharts. However, among these models, only DeepSeek explicitly stated that its in-house training datasets included flowchart-related VQA. This highlights a potential area for future research and development, suggesting that incorporating more flowchart-specific data may enhance model training and performance.

In conclusion, there is a notable gap in resources specifically tailored for flowchart comprehension. Existing datasets and methods primarily focus on general scientific figures without addressing the unique complexities of flowcharts. Our FlowLearn dataset fills this gap by providing detailed annotations for flowcharts and evaluating LVLMs' ability to interpret these diagrams. By enhancing LVLMs' understanding of flowcharts, our work extends existing knowledge and introduces crucial resources for research aimed at enhancing automated visual reasoning and comprehension.

\begin{table*}[]
\resizebox{\textwidth}{!}{%
\begin{tabular}{ll|l|l}
\toprule
\multicolumn{2}{l|}{\textbf{Task}} &
  \multicolumn{1}{l|}{\textbf{Simulated Flowchart}} &
  \multicolumn{1}{l}{\textbf{Scientific Flowchart}} \\
  \hline
 & &
\begin{minipage}{.35\textwidth}
      \includegraphics[width=0.7\linewidth]{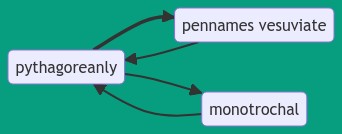}
\end{minipage} 
    &
\begin{minipage}{.35 \textwidth}
      \includegraphics[width=0.5\linewidth]{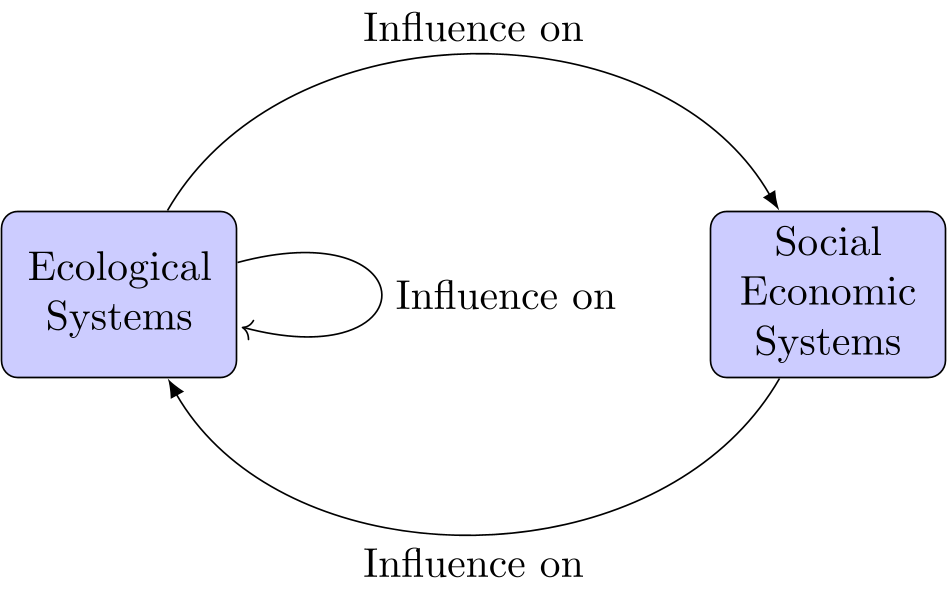}
    \end{minipage}  
   \\\hline
\multirow{3}{*}{\textbf{OCR}} &
  Prompt &
  \multicolumn{2}{c}{\begin{tabular}[c]{@{}c@{}}\textit{A flowchart will be provided where a red box is drawn around the text node of interest.  Answer with} \\ \textit{the text inside the red box. Ensure that the transcription is precise, reflecting the exact letters.}\end{tabular}} \\
 &
  Question &
    \begin{minipage}{.3\textwidth}
      \includegraphics[width=0.4\linewidth]{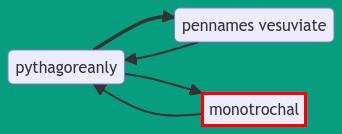}
    \end{minipage} 
    &
  \begin{minipage}{.3\textwidth}
      \includegraphics[width=0.35\linewidth]{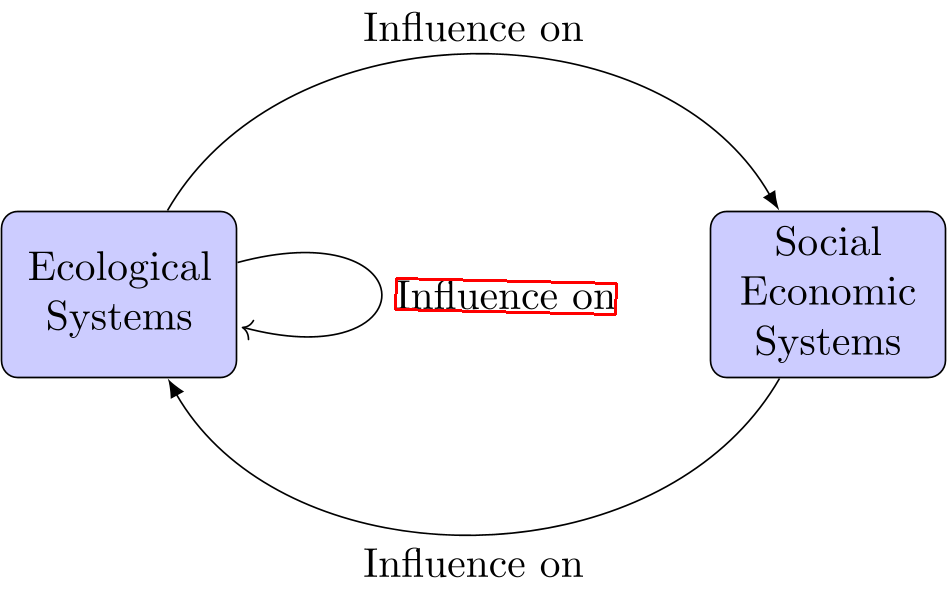}
    \end{minipage}  \\
  
 &
  Answer &
  monotrochal &
  Influence on \\ \hline
\multirow{3}{*}{\textbf{True/False}} &
  Prompt &
  \begin{tabular}[c]{@{}l@{}}\textit{The given image is a \underline{simulated flowchart}. Based on the} \\ \textit{process outlined in the flowchart, determine the correctness }  \\ \textit{of the given statement. Answer with either "true" or "false".}\end{tabular} &
  \begin{tabular}[c]{@{}l@{}}\textit{The given image is a \underline{flowchart extracted from a scientific} }\\\textit{\underline{literature}. Based on the process outlined in the flowchart, } \\ \textit{determine the correctness of the given statement.} \\\textit{Answer with either "true" or "false". }\end{tabular} \\
 &
  Question &
  \begin{tabular}[c]{@{}l@{}}There is an arrow between pennames vesuviate and monotrochal.\end{tabular} &
  Ecological Systems can be influenced by itself. \\
 &
  Answer &
  FALSE &
  TRUE \\ \hline
\multirow{2}{*}{\textbf{Description}} &
  Prompt &
  \multicolumn{2}{c}{\begin{tabular}[c]{@{}c@{}}\textit{The image contains a flowchart. Generate the description of the flowchart, reflecting the text nodes and arrows as depicted.}\end{tabular}} \\
 &
  Answer &
  \begin{tabular}[c]{@{}l@{}}pythagoreanly points to pennames vesuviate. pythagoreanly \\  points to monotrochal. monotrochal points to pythagoreanly. \\ pennames vesuviate points to pythagoreanly.\end{tabular} &
  \begin{tabular}[c]{@{}l@{}}Figure 7: Influence matrix schematic graph, based on \\ {[}5, Figure 5{]}\end{tabular} \\  
  \bottomrule
\end{tabular}%
}
\caption{Common VQA tasks across both the Scientific and Simulated subsets of the FlowLearn Dataset.}
\label{tab:common_vqa}
\end{table*}

\begin{table}[]
\resizebox{\linewidth}{!}{%
\begin{tabular}{ll|l}
\toprule
\multicolumn{2}{l|}{\textbf{Task}} & \textbf{Simulated Flowchart} \\ \hline
\multirow{2}{*}{\textbf{Mermaid Code}} &
  Prompt &
  \begin{tabular}[c]{@{}l@{}}\textit{The image contains a flowchart. Generate the} \\ \textit{Mermaid code to represent the flowchart,} \\ \textit{reflecting the text nodes and arrows as depicted.}\end{tabular} \\
 &
  Answer &
  \begin{tabular}[c]{@{}l@{}}\`{}\`{}\`{}mermaid\\ flowchart LR\\ entity0(pythagoreanly)\\ entity1(monotrochal)\\ entity2(pennames vesuviate)\\ entity0 ==\textgreater entity2\\ entity0 --\textgreater entity1\\ entity1 --\textgreater entity0\\ entity2 --\textgreater entity0\\ \`{}\`{}\`{}\end{tabular} \\ \hline
\multirow{2}{*}{\textbf{Num. of Nodes}} &
  Prompt &
  \begin{tabular}[c]{@{}l@{}}\textit{The given image contains a simulated flowchart.} \\ \textit{You should find all \underline{text nodes} and determine the} \\ \textit{total number of \underline{text nodes} in the flowchart.} \\ \textit{Answer the question with a number.}\end{tabular} \\
          & Answer         & 3                   \\ \hline
\multirow{2}{*}{\textbf{Num. of Arrows}} &
  Prompt &
  \begin{tabular}[c]{@{}l@{}}\textit{The given image contains a simulated flowchart.} \\ \textit{You should find all \underline{arrows} and determine the} \\ \textit{total number of \underline{arrows} in the flowchart.} \\ \textit{Answer the question with a number.}\end{tabular} \\
          & Answer         & 4                   \\ \bottomrule
\end{tabular}%
}
\caption{VQA tasks unique to the Simulated Flowcharts subset of the FlowLearn Dataset.}
\label{tab:vqa_sim}
\end{table}



\section{FlowLearn Dataset}
To address the scarcity of resources for flowchart comprehension, we introduce the FlowLearn Dataset. This dataset comprises two distinct subsets: Scientific Flowcharts and Simulated Flowcharts. An overview of the FlowLearn dataset, illustrating its components, is depicted in Figure \ref{fig:meta}. Table \ref{tab:common_vqa} details the common VQA tasks applicable to both subsets, while Table \ref{tab:vqa_sim} lists the VQA tasks that are unique to the Simulated Flowcharts subset.

\subsection{Scientific Flowchart Dataset}    

The Scientific Flowcharts Dataset comprises a 
large collection of flowchart images extracted from scholarly articles across diverse scientific domains. This dataset serves as a crucial resource for enhancing visual comprehension of scientific content.

We initiated our dataset creation by downloading 27,000 scientific articles from ArXiv. Using PDFFigures 2.0 \cite{pdffigures2}, we extracted figures and related metadata. Additional metadata were parsed by the SciPDF Parser\footnote{https://github.com/titipata/scipdf\_parser}, which utilizes GROBID\footnote{https://github.com/kermitt2/grobid} for parsing PDFs.

Our selection process involved rule-based filtering combined with manual verification to identify flowcharts.
We selected figures based on keywords relevant to flowcharts in captions, such as ``illustration'', ``flowchart'', ``model'', ``step'', ``overall'', and ``graphical representation''. Figures with unrelated keywords like "normalized" and "plot" were omitted. This meticulous curation yielded 3,858 flowcharts from 2,674 documents, focusing on images that prominently feature arrows, indicative of flowchart structures.


Each 
flowchart in our dataset is accompanied by comprehensive metadata, exemplified in Figure~\ref{fig:meta}. The \textbf{Scientific Paper Meta} includes parsed text from the source articles, along with related information such as authors and titles. The \textbf{Figure Meta} encompasses the figure caption and the in-text reference of the figure. Additionally, we annotated all text appearing within each flowchart using PaddleOCR \cite{paddleocr}. These annotations support various subtasks crucial to flowchart comprehension, including OCR and flowchart description.


\subsection{Simulated Flowcharts}\label{sec:sim_content}
Recognizing that understanding flowcharts goes beyond caption generation, we developed the Simulated Flowcharts subset to enhance comprehension of diagrammatic components like arrows and nodes, which can be labor-intensive to annotate in scientific diagrams.

This subset was generated using Mermaid\footnote{https://mermaid.js.org/}, a JavaScript tool that translates Markdown-inspired text definitions into flowcharts. Sample Mermaid code can be seen in Figure \ref{fig:meta} and Table \ref{tab:vqa_sim}. We utilized Python scripts to introduce variability in the flowchart definitions in terms of the following aspects:
\textbf{1) Nodes:} Each flowchart contains between 3 to 10 nodes, with node text consisting of randomized English words.
\textbf{2) Links:} The number of links between nodes is randomized, with all nodes connected by at least one link, mimicking real-world flowchart structures. We randomize the type of arrow links between nodes, including solid lines, bold lines, or dashed lines.
\textbf{3) Background Color and Flowchart Orientation:} Background colors are randomly generated in hexadecimal format. The orientation of the flowcharts is randomized, encompassing all available options in the Mermaid syntax.

We generated a total of 10,000 samples. Each sample includes: \textbf{1) Flowchart Images:} Available in JPEG and SVG formats.
\textbf{2)Mermaid Code:} Provided for each sample to facilitate programmatic understanding and manipulation of the flowchart structure.
\textbf{3)Visual Component Annotations:} Detailed annotations are provided, which include the node text and the precise locations of text nodes, arrowheads, and tails, all derived from SVG. These annotations are crucial for tasks such as object detection and structural analysis, enabling a deeper understanding of the flowchart components.

The generation script provides fine-grained control over the creation of simulated flowchart samples, enabling integrated training and experimentation for a wide range of applications.

\subsection{Visual Question Answering}\label{sec:VQA}
To evaluate the flowchart understanding capabilities using the FlowLearn dataset, we developed tailored 
VQA question-answer pairs for each tested flowchart. Examples of prompts, questions and answers for each task are detailed in Table \ref{tab:common_vqa} and Table \ref{tab:vqa_sim}. We have ensured that all prompts are elaborately detailed based on findings from VL-ICL \cite{vlicl}, which demonstrated that more detailed prompts significantly enhance VQA performance compared to shorter ones. Our own experiments confirm this finding, as we observed that detailed prompts consistently outperform shorter ones in eliciting accurate responses from models.

The common VQA tasks for both subsets include:

\textbf{OCR:} We randomly place a red box over one of the annotated texts within the flowchart and prompt models to identify and return the enclosed words.

\textbf{True/False:} We generate statements related to the flowchart and query the model to determine their veracity. For \textit{Scientific Flowcharts}, we initially create two accurate statements using sentences from the figure caption, subsequently verified by annotators for their correctness based on the flowchart. In cases with insufficient caption data, annotators generate additional statements  relate to the flowchart. For false statements, annotators alter a few words in a true statement to reverse its meaning, ensuring the vocabulary remains consistent with the original author’s style. This process yields one true and one false statement for each tested scientific flowchart. 

For \textit{Simulated Flowcharts}, we use predefined templates to create True and False statements, such as: "An arrow exists between node '\{a\}' and node '\{b\}'" and "An arrow points from node '\{a\}' to node '\{b\}'." where \{a\} and \{b\} are placeholders for node texts identified in Visual Component Annotations (Section \ref{sec:sim_content}).

\textbf{Description:} We prompt models to generate descriptions for the flowcharts. For \textit{scientific flowcharts}, the reference answers are derived from their captions; for \textit{simulated flowcharts}, reference answers are generated by converting mermaid code to sentences using templates, ``\{a\} points to \{b\}.''

Additionally, the Simulated Flowcharts includes 3 unique tasks:

\textbf{Mermaid Code}: Models are tasked with generating Mermaid code that represents the flowchart. This task assesses the model's ability to comprehensively recognize flowchart components, including text nodes and arrows.

\textbf{Number of Nodes and Arrows}: Models answer questions regarding the count of text nodes and arrows present in the flowchart. This task offers a quantitative measure of the model's comprehension, though it is less comprehensive than the Mermaid Code task.

\section{Experiment Setups}
In this section, we detail the experimental setup used to assess the capabilities of various Large Vision-Language Models (LVLMs) using the FlowLearn Dataset. Our primary objective is to evaluate how effectively these models comprehend and interpret flowcharts from both the Scientific and Simulated subsets. We have implemented all VQA tasks outlined in Section \ref{sec:VQA}, which probe various facets of flowchart comprehension—from fundamental text recognition to more comprehensive overall understanding.

\subsection{Models}
We selected LVLMs for evaluation based on their rankings in the OpenCampass multi-modal leaderboard as of April 2024. Access to some models was facilitated through APIs, including Step-1V-32K, GPT-4V, Gemini-Pro-Vision, and Claude-3-Opus-20240229. Additional models assessed in our study were LLaVA-V1.6-Vicuna-34B, InternLM-XComposer2-VL-7B, Qwen-VL-Chat from 2024/01/25, and DeepSeek-VL-7B-chat. Our selection strategy aimed to choose the best model available from each top-ranked model family, such as selecting Claude-3-Opus from the Claude series. 

\subsection{Evaluation Metrics}
To evaluate the performance of the models, we categorized the VQA tasks into three groups, each assessed by tailored evaluation metrics:

\textbf{Accuracy:} 
We measure the accuracy for tasks including OCR, True/False Statements, Number of Nodes, and Number of Arrows. This metric is straightforward and evaluates whether the responses are correct or incorrect based on the ground truth. Specifically for True/False Statements, we calculated average accuracy separately for the true and false subsets, and an overall average accuracy to provide a comprehensive view of model performance.

\textbf{Similarity:} 
For description tasks, we assess the closeness of model-generated descriptions to reference descriptions using four similarity metrics: BLEU~\cite{bleu}, ROUGE-L~\cite{rouge}, BERT score~\cite{bertscore} and Sentence Transformer Similarity (SBERT)~\cite{sentencebert}. The BERT score utilizes pre-trained BERT embeddings to assess semantic coherence through cosine similarity of matched words. Similarly, SBERT converts both response and reference sentences into embeddings with the `all-MiniLM-L6-v2' model, using cosine similarity to quantitatively gauge how closely the generated text matches the target description. We also calculate median word count of responses.


\textbf{Mermaid Code Generation:} 
We developed two sets of metrics specifically tailored for evaluating the correctness of generated Mermaid code: 
\begin{itemize}
\item \textit{Node-Level Evaluation:} This metric checks if all nodes present in the ground truth are included in the model's response. Each node is only considered correct if it exactly matches the spelling in the ground truth.  \item \textit{Link-Level Evaluation:} This metric assesses the generated response includes all the links present in the ground truth. A link is deemed correct if both the start and end nodes are accurately predicted, regardless of the arrow type. We also permit some syntactical flexibility in how node descriptions are expressed, allowing the use of either the node variable name or the node text. 
\end{itemize}

For both evaluation levels, we compute F1-score, precision, and recall for each sample and average these metrics across all samples.

\subsection{Response Parsing}
Given the variability in how LVLMs generate responses, which may not always exactly match the ground truth even when correct, we have developed specific rules to parse and evaluate the responses:

 \textbf{OCR:} A prediction is deemed correct if it includes the exact phrase from the ground truth.
 
 \textbf{True/False Statements:} The response is assessed for the presence of 'true' or 'false', case-insensitively. Responses lacking these tokens or containing both are marked as incorrect.
 
 \textbf{Number of Nodes or Arrows:} We extract the first numeric token in the response, also converting English words representing numbers into numeric tokens. If no such token appears, the response is marked as incorrect.
 
 \textbf{Mermaid Code Prediction:} We focus on statements encapsulated within triple backticks (\`{}\`{}\`{}) in model responses. From these, we extract nodes and links according to the Mermaid syntax rules.


\subsection{Settings}
For our evaluations, we utilized the testing subset of the FlowLearn dataset, which included assessments of 500 scientific flowcharts and 2,000 simulated flowcharts. Due to cost constraints and API limitations, we limited our evaluations to 100 samples per task for Claude-3-Opus, GPT-4V, and Step-1V. All other evaluations were conducted using an NVIDIA A100 80GB GPU.

We opted for few-shot prompting as our evaluation strategy to align the output of the 
LVLMs more closely with the ground truth. According to \citet{vlicl}, few-shot prompting, particularly with 2-shot samples, generally yields the most significant performance improvement in general vision-language VQA tasks across various LVLMs. Additionally, using 2-shot samples provides a balanced approach for evaluating True/False statements, as it allows an equal representation of both true and false scenarios within the prompts. This method ensures that the models are not biased toward one answer type over the other, facilitating a more accurate and fair assessment of model capabilities.

For consistency, we employ the prompt format shown in Table \ref{tab:prompt} for evaluation. 

\vspace{-5pt}
\begin{table}[]
\centering
\begin{tabular}{|
>{\columncolor[HTML]{C0C0C0}}l |}
\hline
\textbf{Prompt:} {[}Task Description{]}                              \\
\textbf{Support Set:} {[}Image{]}{[}Question{]}{[}Answer{]} (2-shot) \\
\textbf{Query:} {[}Image{]}{[}Question{]}                            \\
\textbf{Prediction:} {[}Answer{]}                                    \\ \hline
\end{tabular}%
\caption{2-Shot prompt format used for evaluation.}
\label{tab:prompt}
\end{table}
\vspace{-10pt}
\section{Experiment Results}
In this section, we present the results from our evaluation of the 
LVLMs across three distinct groups of 
VQA tasks within the FlowLearn dataset. Each task group was designed to test different aspects of model performance using specialized evaluation metrics. For a focused review of performance across a limited subset of 100 samples involving all models and all tasks, please refer to Section 2 of the Supplementary Materials. The findings there align closely with the results discussed here. Sample model responses to all VQA tasks are shown in Section 3 and 4 of the Supplementary Materials.

\begin{table*}[]
\centering
\begin{tabular}{lrrrrrrrrrr}
\toprule
\multicolumn{2}{c|}{Task} &
  Claude\ensuremath{\dagger} &
  GPT4V\ensuremath{\dagger} &
  Step-1V\ensuremath{\dagger} &
  \begin{tabular}[c]{@{}r@{}}Gemini\\ ProVision\end{tabular} &
  \begin{tabular}[c]{@{}r@{}}InternLM\\ -XComposer2-VL\end{tabular} &
  \begin{tabular}[c]{@{}r@{}}LLaVA\\ 16-34B\end{tabular} &
  \begin{tabular}[c]{@{}r@{}}Qwen-VL\\ -chat\end{tabular} &
  \begin{tabular}[c]{@{}r@{}}DeepSeek\\ -VL-7B-chat\end{tabular} \\ \hline
  
\multicolumn{10}{c}{\cellcolor[HTML]{F3F3F3}Scientific Flowchart}                                                                              \\ \hline
\multicolumn{2}{r|}{OCR}         & 0.44                & 0.51          & \textbf{0.66} & {\ul 0.43} & 0.06                & 0.01 & 0.08 & 0.05 \\
   & \multicolumn{1}{r|}{TRUE}   & 0.69                & 0.63          & \textbf{0.9}  & 0.55       & {\ul 0.89}          & 0.15 & 0.18 & 0.86 \\
   & \multicolumn{1}{r|}{FALSE}  & 0.53                & \textbf{0.74} & 0.36          & {\ul 0.7}  & 0.28                & 0    & 0    & 0.21 \\
\multirow{-3}{*}{Statements} &
  \multicolumn{1}{r|}{Average} &
  0.61 &
  \textbf{0.68} &
  0.63 &
  {\ul 0.62} &
  0.59 &
  0.08 &
  0.09 &
  0.53 \\ \hline
\multicolumn{10}{c}{\cellcolor[HTML]{EFEFEF}Simulated Flowchart}                                                                               \\ \hline
\multicolumn{2}{r|}{OCR}         &  \textbf{0.83} & 0.75          & 0.71          & {\ul 0.69}       & 0.18                & 0    & 0.58 & 0.23 \\
\multicolumn{2}{r|}{Num. Nodes}  & 0.52                & \textbf{0.58} & 0.31          & 0.02       & 0.15                & 0.12 & 0.4  & 0.22 \\
\multicolumn{2}{r|}{Num. Arrows} & 0.23                & \textbf{0.26} & \textbf{0.26} & 0.09       & 0.12                & 0.1  & 0.2  & 0.15 \\ \hline
   & \multicolumn{1}{r|}{TRUE}   & 0.42                & 0.28          & 0.62          & 0.69       & {\ul \textbf{0.82}} & 0.05 & 0.34 & 0.61 \\
   & \multicolumn{1}{r|}{FALSE}  & 0.71                & 0.77          & 0.72          & 0.52       & 0.5                 & 0.01 & 0    & 0.56 \\
\multirow{-3}{*}{Statements: Between AB} &
  \multicolumn{1}{r|}{Average} &
  0.56 &
  0.52 &
  \textbf{0.67} &
  0.61 &
  {\ul 0.66} &
  0.03 &
  0.17 &
  0.59 \\ \hline
   & \multicolumn{1}{r|}{TRUE}   & 0.18                & 0.15          & 0.41          & 0.16       & {\ul \textbf{0.85}} & 0.11 & 0.04 & 0.55 \\
   & \multicolumn{1}{r|}{FALSE}  & 0.49                & 0.61          & 0.5           & 0.63       & 0.49                & 0.03 & 0.01 & 0.58 \\
\multirow{-3}{*}{Statements: A to B} &
  \multicolumn{1}{r|}{Average} &
  0.34 &
  0.38 &
  0.45 &
  0.4 &
  {\ul \textbf{0.67}} &
  0.07 &
  0.03 &
  0.57

  \\ \bottomrule
\end{tabular}%

\caption{Experiment results for accuracy tasks. Models\ensuremath{\dagger} are evaluated on a subset of the evaluation set. Regardless of evaluation size, the best-performing model is \textbf{bolded}. The best-performing model among those evaluated on the full set is \underline{underlined}.}
\label{tab:exp-acc}
\end{table*}

\subsection{Accuracy Tasks} 
The first group of tasks evaluates the accuracy of the LVLMs in responding to queries that require precise, binary, or short phrase answers. These tasks are foundational for assessing flowchart comprehension. The performance of each model on these accuracy tasks is summarized in Table \ref{tab:exp-acc}, leading to several key observations:

1) \textbf{No clear winner across all accuracy tasks.} For scientific flowcharts, Gemini-Pro-Vision showed the strongest performance on the full test set. However, on smaller subsets, GPT-4V and Step-1V also demonstrated strong performances. For simulated flowcharts, on the full test set, InternLM excelled in True/False statements, Gemini-Pro in OCR tasks, and Qwen-VL in counting nodes and arrows.

2) \textbf{Irrelevant model responses.} Although most models generally produced task-related responses, irrelevant responses were still observed. For True/False tasks, Qwen-VL and LLaVA often scored close to zero, indicating a lack of 'true' or 'false' tokens in its responses. 

3) \textbf{Challenges in counting nodes and arrows.} Counting tasks, which require comprehensive image understanding rather than partial recognition, proved difficult for most models, leading to lower average scores. Notably, despite its underperformance in other areas, Qwen-VL's results were comparatively better in these tasks.

\begin{table*}[]
\centering
\begin{tabular}{rr|rrrrrrrr}
\toprule
\multicolumn{2}{r|}{Evaluation Metrics} &
  Claude\ensuremath{\dagger} &
  GPT4V\ensuremath{\dagger} &
  Step-1V\ensuremath{\dagger} &
  \begin{tabular}[c]{@{}r@{}}Gemini\\ ProVision\end{tabular} &
  \begin{tabular}[c]{@{}r@{}}InternLM\\ -XComposer2-VL\end{tabular} &
  \begin{tabular}[c]{@{}r@{}}LLaVA\\ 16-34B\end{tabular} &
  \begin{tabular}[c]{@{}r@{}}Qwen-VL\\ -chat\end{tabular} &
  \begin{tabular}[c]{@{}r@{}}DeepSeek\\ -VL-7B-chat\end{tabular} \\ \hline

\multicolumn{10}{c}{\cellcolor[HTML]{F3F3F3}Scientific Flowchart}                                                                                     \\ \hline
\multicolumn{2}{r|}{Median Word Count}      & 98            & 97   & 65   & 190           & 101  & 19   & 354  & 9             \\
\multicolumn{2}{r|}{BLEU}                        & 0.47          & 0.43 & 0.33 & {\ul \textbf{0.49}} & 0.05 & 0    & 0.02       & 0.13                \\
\multicolumn{2}{r|}{ROUGE-L}                     & 0.13          & 0.13 & 0.12 & 0.09                & 0.1  & 0.11 & 0.07       & {\ul \textbf{0.17}} \\
\multicolumn{2}{r|}{BERTScore-F1}                & 0.84          & 0.83 & 0.84 & 0.83                & 0.83 & 0.81 & 0.79       & {\ul \textbf{0.86}} \\
\multicolumn{2}{r|}{Sent.Transformer Similarity} & \textbf{0.49} & 0.46 & 0.42 & 0.30                & 0.34 & 0.25 & {\ul 0.38} & 0.36                \\ \hline
\multicolumn{10}{c}{\cellcolor[HTML]{EFEFEF}Simulated Flowchart}                                                                                      \\ \hline
\multicolumn{2}{r|}{Median Word Count}      & 28            & 34   & 35   & 33            & 58   & 19   & 200  & 32            \\
\multicolumn{2}{r|}{BLEU}                        & \textbf{0.02} & 0.01 & 0.01 & 0.01                & 0.01 & 0    & 0          & {\ul \textbf{0.02}} \\
\multicolumn{2}{r|}{ROUGE-L}                     & 0.54          & 0.52 & 0.52 & {\ul \textbf{0.56}} & 0.19 & 0    & 0.13       & 0.41                \\
\multicolumn{2}{r|}{BERTScore-F1}                & 0.90          & 0.90 & 0.89 & {\ul \textbf{0.92}} & 0.82 & 0.77 & 0.80       & 0.87                \\
\multicolumn{2}{r|}{Sent.Transformer Similarity} & 0.84          & 0.84 & 0.84 & {\ul \textbf{0.88}} & 0.41 & 0.18 & 0.51       & 0.71                

\\ \bottomrule 
\end{tabular}%
\caption{Experiment results for Flowchart Description task. Models\ensuremath{\dagger} are evaluated on a subset of the evaluation set. Regardless of evaluation size, the best-performing model is \textbf{bolded}. The best-performing model among those evaluated on the full set is \underline{underlined}.}
\label{tab:exp-des}
\end{table*}

\begin{table}[]
\begin{tabular}{l}
\toprule
\begin{minipage}{0.95\linewidth}
\includegraphics[width=0.95\linewidth]{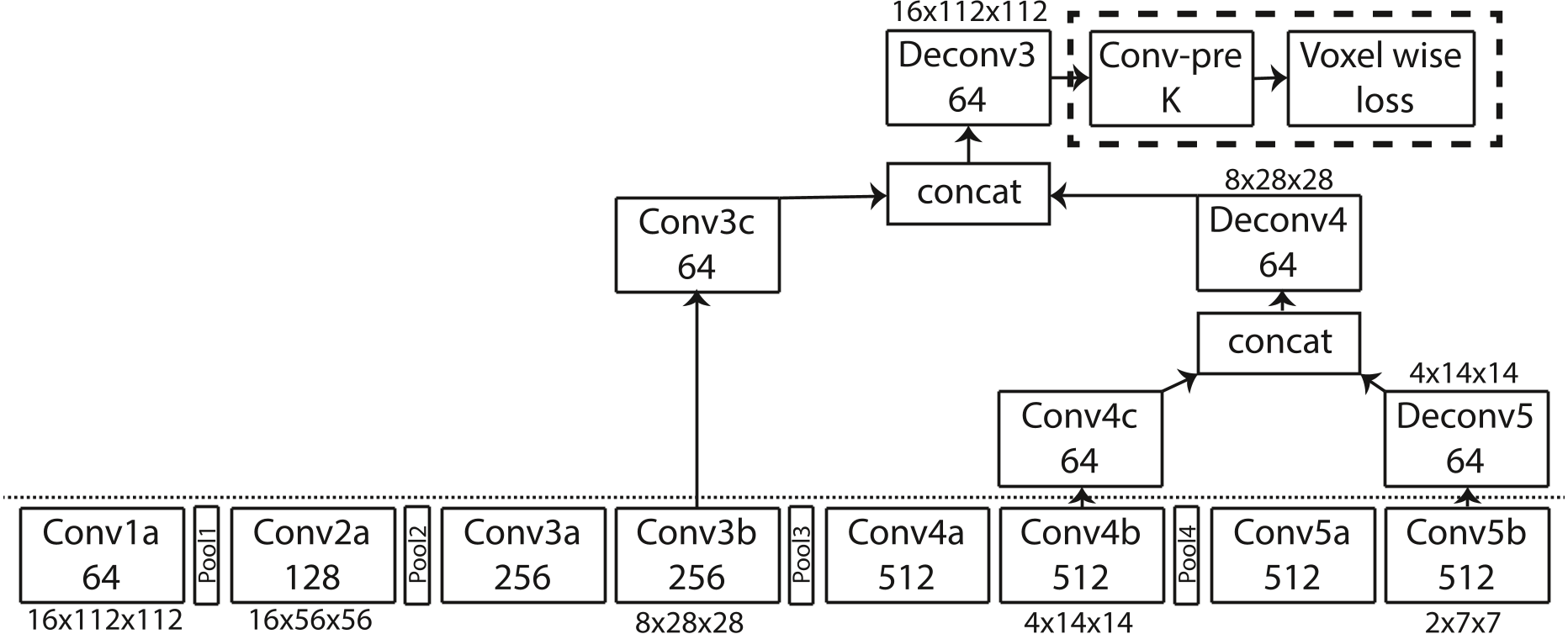}
\end{minipage}                                                                                                              \\
\begin{tabular}[c]{@{}l@{}}
 The image depicts a neural network architecture for processing  \\
volumetric data, possibly for a segmentation task. It features an \textcolor{red}{encoder}-\\ \textcolor{red}{decoder structure with skip connections.} Here is the flow:\\
\textcolor{red}{1. The network inputs a volume of 16x112x112x1}. \\
2. It goes through four convolutional layers with the following \\ specifications: \\
- Conv1a: 64 feature maps, output size 16x112x112\\
- Conv2a: 128 feature maps, output size 16x56x56\\
- Conv3a: 256 feature maps, output size 8x28x28 \\
- Conv4a: 512 feature maps, output size 4x14x14   \\
3. There's another convolutional layer Conv5a with 512 feature \\
maps reducing the volume size to 2x7x7.\\
... \textit{(cropped) }... \\
Lastly, there's a Conv-pre layer (size not specified) \textcolor{red}{followed by} \\
\textcolor{red}{Bounding Box (BB) Regression to predict 'K' classes, }and a voxel-wise  \\
loss is computed at the final output.\\
Note: The sizes of the feature maps in the decoder part are not explicitly\\ stated, but they are likely to match the corresponding encoder layer sizes \\due to skip connections. The dotted lines indicate feedback of loss or\\ gradients back through the network during training.
\end{tabular}\\ \bottomrule
\end{tabular}%
\caption{Example GPT-4V's response (cropped) to the Flowchart-to-Caption tasks on Scientific Flowcharts from the FlowLearn Dataset. Incorrect parts are highlighted in \textcolor{red}{red}.}
\label{tab:exp-gpt4v-des}
\end{table}

\subsection{Similarity Tasks (Description)}
The second group of tasks assesses LVLMs' ability to generate accurate descriptions of flowcharts, with detailed performance data for each model presented in Table \ref{tab:exp-des}. 

For scientific flowcharts, DeepSeek and Gemini achieved the highest scores on most metrics. Qwen-VL produced the longest responses, while DeepSeek provided the shortest. We also reviewed responses from the widely-used GPT4V. Out of 100 samples, only 24 were error-free. Analysis at the sentence level yielded 597 sentences with a 59\% accuracy rate, meaning 41\% of sentences contained errors. Sentences were marked as incorrect only if evidence from the image contradicted the text. Supplementary material includes examples. We identified several common trends:

1) Generally, models provided satisfactory responses, accurately inferring full names from acronyms and offering reasonable academic interpretations of the depicted processes. 

2) Models handled complex images effectively, including those with high resolutions and multiple parts. 

3) Longer descriptions were more error-prone, whereas shorter ones, while accurate, lacked comprehensive coverage, explaining the lower sample-wise compared to sentence-wise accuracy.


4) In many cases, models produced logical but inaccurate descriptions. Examples of this are shown in Table~\ref{tab:exp-gpt4v-des}.


For simulated flowcharts, Gemini outperformed other models across most metrics, typically scoring higher than for scientific flowcharts. This discrepancy likely stems from the structured, template-generated reference answers for simulated flowcharts, contrasted with the varied language and additional contextual information in scientific flowchart captions. 

\begin{table*}[]
\centering
\begin{tabular}{rr|rrrrrrrrr}
\toprule
\multicolumn{2}{c|}{Evaluation Metric} &
  Claude\ensuremath{\dagger} &
  GPT4V\ensuremath{\dagger} &
  Step-1V\ensuremath{\dagger} &
  \begin{tabular}[c]{@{}r@{}}Gemini\\ ProVision\end{tabular} &
  \begin{tabular}[c]{@{}r@{}}Gemini\\ ProVision (CoT)\end{tabular} &
  \begin{tabular}[c]{@{}r@{}}InternLM\\ -XComposer2-VL\end{tabular} &
  \begin{tabular}[c]{@{}r@{}}LLaVA\\ 16-34B\end{tabular} &
  \begin{tabular}[c]{@{}r@{}}Qwen-VL\\ -chat\end{tabular} &
  \begin{tabular}[c]{@{}r@{}}DeepSeek\\ -VL-7B-chat\end{tabular} \\ \hline
\multirow{3}{*}{Link} & Precision & \textbf{0.35} & 0.23 & 0.14 & {\ul 0.26} & 0.25 & 0.01 & 0 & 0.02 & 0.05 \\
                      & Recall    & \textbf{0.26} & 0.22 & 0.15 & {\ul 0.25} & 0.24 & 0.02 & 0 & 0.02 & 0.04 \\
                      & F1        & \textbf{0.3}  & 0.22 & 0.14 & {\ul 0.25} & 0.25 & 0.02 & 0 & 0.02 & 0.04 \\ \hline
\multirow{3}{*}{Node} & Precision & \textbf{0.94} & 0.72 & 0.68 & {\ul 0.75} & 0.71 & 0.09 & 0 & 0.06 & 0.29 \\
                      & Recall    & \textbf{0.95} & 0.73 & 0.68 & {\ul 0.75} & 0.71 & 0.16 & 0 & 0.08 & 0.29 \\
                      & F1        & \textbf{0.94} & 0.72 & 0.68 & {\ul 0.75} & 0.71 & 0.12 & 0 & 0.07 & 0.28
                      \\
\bottomrule
\end{tabular}%
\caption{Results for Flowchart-to-Mermaid on Simulated Flowcharts. Models\ensuremath{\dagger} are evaluated on a subset of the evaluation set. Regardless of evaluation size, the best-performing model is \textbf{bolded}. The best-performing model among those evaluated on the full set is \underline{underlined}.}
\label{tab:exp-mermaid}
\end{table*}

\subsection{Mermaid Code Task} \label{sec:expr-mermaid}
This task assesses the comprehensive ability of LVLMs to encapsulate their understanding of a flowchart in a code format, summarizing aspects such as OCR, counting nodes and arrows, and recognizing relationships between nodes. The performance of each model on the Mermaid Code task for simulated flowcharts is summarized in Table \ref{tab:exp-mermaid}. In evaluations on the full dataset, Gemini achieved the highest scores across all metrics. On a smaller evaluation subset, Claude demonstrated superior performance, particularly excelling in node-level prediction with an F1 score of 94\%.

Challenges were notable in models like InternLM, LLaVA, Qwen-VL, and DeepSeek, all of which recorded scores close to zero. Several issues were identified during the evaluation of their outputs: 

\textbf{Syntax Compliance:} These models did not adhere to the proper syntax of Mermaid code, failing to correct their outputs even after 2-shot prompting designed to teach them the correct code format.

\textbf{Node Recognition:} Disregarding syntax issues, these models still struggled to accurately predict correct nodes. The node-level evaluation, which also indirectly assesses models' OCR capabilities by checking for the presence of all node text in the predictions, reflected poor performance. This aligns with results from Table \ref{tab:exp-acc}, where these models underperformed in OCR tasks that required text detection within specified areas.

Link-level predictions, which depend on accurate node-level results, consider a prediction correct only if the start and end nodes and the direction of the link are identified accurately. Hence, scores for link-level evaluations generally fall below those for node-level evaluations. Claude, which scored highly at the node level, encountered many challenges with link prediction, achieving only a 30\% F1 score for link-level accuracy. This highlights the difficulty models face in understanding complex relationships within flowcharts.

\subsection{Ablation Study on Chain-of-Thought}
For complex tasks such as converting a flowchart into Mermaid code, a methodical approach can be beneficial. This process typically involves several sequential steps: initially detecting text nodes, then recognizing the links between them, and finally compiling these information into a standardized format, such as Mermaid code. Given the multi-step nature of this task, we hypothesized that introducing a chain-of-thought (CoT) process could potentially enhance model performance. Consequently, we conducted an experimental ablation study on the simulated subset using Gemini-Pro-Vision, a model that incurs no querying cost and can be evaluated on the full test set. Notably, this model has shown the best performance on the Mermaid code task (Section \ref{sec:expr-mermaid}).

\begin{table}[]
\centering
\begin{tabular}{|
>{\columncolor[HTML]{C0C0C0}}l |}
\hline
First, the flowchart includes the following nodes: \textbf{\{1\}} \\
Then, it contains the following edges: \textbf{\{2\}} \\
Finally, the Mermaid code for the flowchart is: \textbf{\{3\}} \\ \hline
\end{tabular}%
\caption{Chain-of-Thought answer template.}
\label{tab:cot}
\end{table}

For this experiment, we modified the 2-shot example answers using a structured template (Table \ref{tab:cot}) that guides the model through a step-by-step reasoning process. In this template, \{1\} is replaced with all text appearing in the simulated flowchart, \{2\} is derived from the flowchart description generated as per the templates described in Section \ref{sec:VQA}, and \{3\} is the corresponding Mermaid code. Additionally, we appended the phrase "Let's think step by step" at the end of the original prompt (as illustrated in Table \ref{tab:vqa_sim}) to further emphasize the sequential reasoning process.

We selected Gemini, the top-performing model for the Mermaid Code Task, for this ablation study.  Surprisingly, as shown in Table \ref{tab:exp-mermaid}, the performance using the CoT approach indicated a slight decrease compared to the original model configuration without it. This unexpected outcome suggests that while the chain-of-thought method is intended to foster clearer and more structured reasoning, it may introduce additional complexities or dependencies that hinder the model's ability to synthesize and process information efficiently. Further analysis and refinement of the chain-of-thought implementation may be necessary to fully realize its potential benefits and address these challenges.

\section{Discussion} 
The initial version of the FlowLearn dataset presents inherent limitations, offering opportunities for future enhancements.

\subsection{Scientific Flowchart Subset}
First, True/False statements are missing for the training set of scientific flowcharts. Annotators generate these statements only for test samples, which is time-consuming, averaging three minutes per pair. Future versions should aim to include these statements for all entries.

Second, the dataset size is currently limited. With fewer than 4,000 images, FlowLearn's scientific flowchart collection is small compared to larger visual-language datasets. While the inclusion of simulated flowcharts helps to mitigate this limitation by broadening the scope of the training data, expanding the collection of scientific flowcharts would be advantageous. 

Third, the descriptive task is limited. The descriptive task for scientific flowcharts is currently evaluated against figure captions. However, the descriptive text for scientific diagrams is often scattered throughout the associated literature, as outlined in Context24\footnote{https://sdproc.org/2024/sharedtasks.html\#context24}: Contextualizing Scientific Figures and Tables. A more robust approach would involve annotators extracting and collating descriptive text from the full text of scientific articles to provide a more comprehensive base for evaluating LVLM-generated descriptions.

\subsection{Simulated Flowchart Subset}
The simulated flowchart subset was designed to augment the scientific subset by offering a more granular evaluation of flowchart comprehension and providing additional training data. Future iterations could improve upon this by incorporating a greater diversity of diagram types, such as state diagrams and quadrant charts, to enrich the dataset further. While FlowLearn currently focuses exclusively on flowcharts, expanding the range of diagram types could enhance its applicability.

\subsection{Model Selection}
Our model selection was biased towards LVLMs due to their broad capabilities and general applicability. However, many task-specific, smaller visual-language models may also be well-suited for these tasks. Future work will explore the potential of these models, which might offer more specialized insights or efficiencies in specific aspects of flowchart comprehension.

\section{Conclusion}
In this study, we introduced and evaluated the FlowLearn dataset, a novel resource aimed at advancing the understanding of flowcharts for visual-language models. Our experiments spanned various tasks, including OCR, True/False assessments, counting nodes and arrows, flowchart description, and generating Mermaid code, across two distinct subsets: scientific and simulated flowcharts.

Our findings demonstrate that while LVLMs are capable of impressive performance on certain tasks, challenges remain. Notably, the models excelled at OCR and True/False statements in certain contexts but struggled with the more complex task of accurately generating Mermaid code from flowcharts. This underscores a broader issue: LVLMs often struggle to fully comprehend the intricate relationships between visual components and to synthesize this information into structured code formats effectively. 

Given the rapid advancements in the fields of LLMs and LVLMs, the FlowLearn dataset is timely and provides valuable insights into a relatively underexplored area. It not only serves as a critical tool for benchmarking and refining these models but also helps illuminate the specific difficulties they encounter with visual reasoning in a structured context. By pushing the boundaries of what LVLMs can understand and achieve, we can bridge the gap between human and machine comprehension of visual and language tasks, paving the way for more intelligent and capable automated systems.



\begin{ack}
This work was supported by the National Science Foundation awards III-2107213, III-2107518, and ITE-2333789. We also thank undergraduate students Eric Reizas and Elle Nguyen at Temple University for their valuable contributions to our project.
\end{ack}



\bibliography{main}
\section{Supplementary Materials}
\input{supplementary_arxiv}
\end{document}

%% file: supplementary_arxiv.tex
\subsection{Preliminary experiment with ACL-Fig}
We conducted a preliminary experiment using flowcharts from ACL-Fig, where flowcharts are categorized as neural networks and architecture diagrams. The dataset includes a total of 208 flowcharts. We queried Gemini-Pro-Vision with a prompt requesting a figure description, using the same prompt shown in the main text. By comparing the generated descriptions to the actual figure captions, we obtained surprisingly poor results: BLEU score of 0.006, ROUGE-L of 0.08, BERTScore-F1 of 0.824, and Sentence Transformer Similarity of 0.40. We also manually evaluated the generated descriptions and found that almost all responses contained at least one error, such as incorrect interpretation of an arrow leading to erroneous causation explanations. Examples of these descriptions are illustrated in Table~\ref{tab:aclfig}.

\begin{table}[]
\resizebox{\linewidth}{!}{%
\begin{tabular}{ll}
\toprule
Image File                                                           & 1996.amta-1.9.pdf-Figure1.png                                                                                                                                      \\
Image                                                                &  \begin{minipage}{0.85 \linewidth}
      \includegraphics[width=0.85\linewidth]{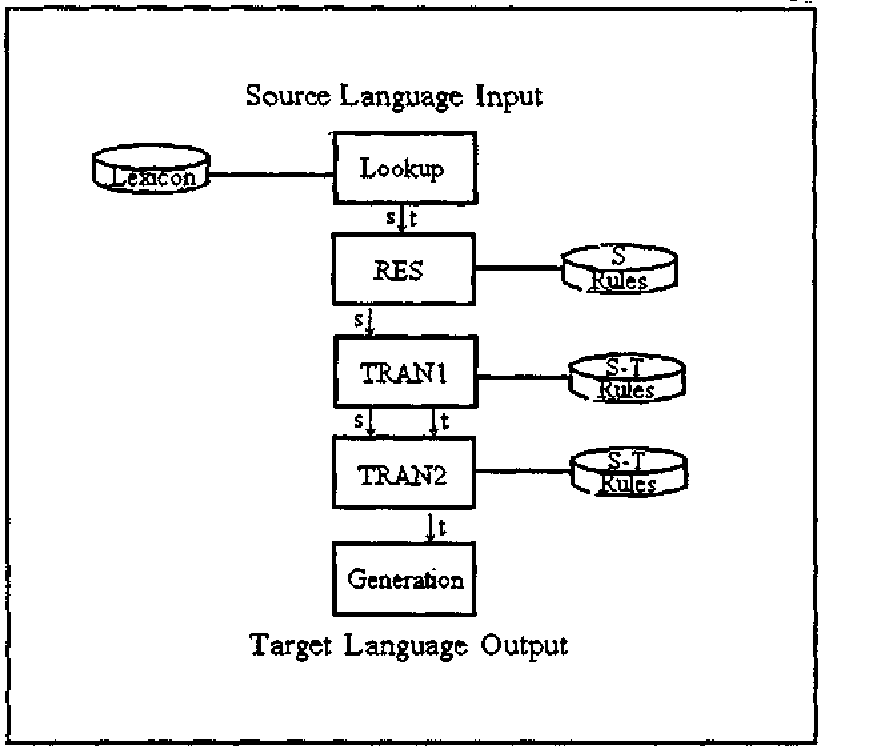}
    \end{minipage}                                                                                                      \\
Caption                                                              &       Figure 1      \\\hline                                                                                 
\begin{tabular}[c]{@{}l@{}}Description\\ from \\ Gemini\end{tabular} & \begin{tabular}[c]{@{}l@{}}The flowchart describes the process of translating a source \\ language into a target language. \\ The process starts with the source language input, \\ which is then looked up in a lexicon. \\ The result of the lookup is then passed to a series of modules, \\ which perform various transformations on the input. \\ The first module is the RES module, \\ which is responsible for resolving anaphora. \\ The output of the RES module is then passed to the S-Rules \\ module, which applies a set of rules to the input. \\ \color{red}{The output of the S-Rules module is then passed to the TRANI} \\ module, which performs a series of transformations on the input. \\ The output of the TRANI module is then passed to the TRAN2 \\ module, which performs a series of transformations on the input. \\ The output of the TRAN2 module is then passed to the \\ Generation module, which generates the target language output.\end{tabular}
\\ \bottomrule
\end{tabular}%
}
\caption{Example description from Gemini-Pro-Vision for a sample in ACL-Fig. Text highlighted in {\color{red}{red}} indicates parts that incorrectly describe the figure.}
\label{tab:aclfig}
\end{table}

\subsection{Experiment Results on a Limited Subset}
Due to cost and API limitations, the models Claude, GPT-4V, and Step-1V were evaluated on a smaller subset of 100 samples. To provide a comprehensive comparison, we present the experiment results for all models across this smaller subset in Table \ref{tab:exp100-acc}, Table \ref{tab:exp100-sim}, and Table \ref{tab:exp100-mermaid}. The findings from this limited subset are consistent with the results discussed in the main text, underscoring the robustness and reliability of our earlier observations.

\begin{table*}[]
\resizebox{\textwidth}{!}{%
\begin{tabular}{rrrrrrrrrr}
\toprule
\multicolumn{2}{r|}{Task} &
  Claude &
  GPT4V &
  Step-1V &
  \begin{tabular}[c]{@{}r@{}}Gemini\\ ProVision\end{tabular} &
  \begin{tabular}[c]{@{}r@{}}InternLM\\ -XComposer2-VL\end{tabular} &
  LLaVA16-34B &
  \begin{tabular}[c]{@{}r@{}}Qwen-VL\\ -chat\end{tabular} &
  \begin{tabular}[c]{@{}r@{}}DeepSeek\\ -VL-7B-chat\end{tabular} \\ \hline
  
\multicolumn{10}{c}{\cellcolor[HTML]{F3F3F3}Scientific Flowchart}                                                           \\ \hline
\multicolumn{2}{r|}{OCR}        & 0.44          & 0.51          & \textbf{0.66} & 0.53 & 0.05          & 0.01 & 0.07 & 0.09 \\ \hline
  & \multicolumn{1}{r|}{TRUE}   & 0.69          & 0.63          & \textbf{0.9}  & 0.63 & 0.84          & 0.14 & 0.17 & 0.82 \\
  & \multicolumn{1}{r|}{FALSE}  & 0.53          & \textbf{0.74} & 0.36          & 0.61 & 0.32          & 0    & 0    & 0.21 \\
\multirow{-3}{*}{Statements}             & \multicolumn{1}{r|}{Average} & 0.61 & \textbf{0.68} & 0.63          & 0.62 & 0.58          & 0.07 & 0.08 & 0.52 \\ \hline
\multicolumn{10}{c}{\cellcolor[HTML]{EFEFEF}Simulated Flowchart}                                                            \\ \hline
\multicolumn{2}{r|}{OCR}        & \textbf{0.83} & 0.75          & 0.71          & 0.64 & 0.19          & 0    & 0.6  & 0.2  \\
\multicolumn{2}{r|}{Num. Nodes} & 0.52          & \textbf{0.58} & 0.31          & 0.04 & 0.15          & 0.08 & 0.42 & 0.22 \\
\multicolumn{2}{r|}{Num. Arrows}                                        & 0.23 & \textbf{0.26} & \textbf{0.26} & 0.07 & 0.07          & 0.07 & 0.19 & 0.12 \\ \hline
  & \multicolumn{1}{r|}{TRUE}   & 0.42          & 0.28          & 0.62          & 0.68 & \textbf{0.79} & 0.26 & 0.34 & 0.74 \\
  & \multicolumn{1}{r|}{FALSE}  & 0.71          & \textbf{0.77} & 0.72          & 0.49 & 0.44          & 0.01 & 0    & 0.5  \\
\multirow{-3}{*}{Statements: Between AB} & \multicolumn{1}{r|}{Average} & 0.56 & 0.52          & \textbf{0.67} & 0.58 & 0.62          & 0.14 & 0.17 & 0.62 \\ \hline
  & \multicolumn{1}{r|}{TRUE}   & 0.18          & 0.15          & 0.41          & 0.13 & \textbf{0.83} & 0.79 & 0.04 & 0.51 \\
  & \multicolumn{1}{r|}{FALSE}  & 0.49          & \textbf{0.61} & 0.5           & 0.53 & 0.4           & 0.33 & 0.02 & 0.52 \\
\multirow{-3}{*}{Statements: A to B}     & \multicolumn{1}{r|}{Average} & 0.34 & 0.38          & 0.45          & 0.33 & \textbf{0.62} & 0.56 & 0.03 & 0.52
  \\ \bottomrule
\end{tabular}%
}
\caption{Experiment results for accuracy tasks across the same 100 samples. The best-performing model is highlighted in \textbf{bolded}.}
\label{tab:exp100-acc}
\end{table*}

\begin{table*}[]
\resizebox{\textwidth}{!}{%
\begin{tabular}{rlrrrrrrrr}
\toprule
\multicolumn{2}{r|}{Evaluation Metrics} &
  Claude &
  GPT4V &
  Step-1V &
  \begin{tabular}[c]{@{}r@{}}Gemini\\ ProVision\end{tabular} &
  \begin{tabular}[c]{@{}r@{}}InternLM\\ -XComposer2-VL\end{tabular} &
  LLaVA16-7B &
  \begin{tabular}[c]{@{}r@{}}Qwen-VL\\ -chat\end{tabular} &
  \begin{tabular}[c]{@{}r@{}}DeepSeek\\ -VL-7B-chat\end{tabular} \\ \hline
  
\multicolumn{10}{c}{\cellcolor[HTML]{F3F3F3}Scientific Flowchart}                                                                   \\ \hline
\multicolumn{2}{r|}{BLEU}                        & 0.02          & 0.01 & 0.01 & 0.01          & 0.01 & 0    & 0    & \textbf{0.03} \\
\multicolumn{2}{r|}{ROUGE-L}                     & 0.13          & 0.13 & 0.12 & 0.1           & 0.1  & 0.11 & 0.07 & \textbf{0.18} \\
\multicolumn{2}{r|}{BERTScore-F1}                & 0.84          & 0.83 & 0.83 & 0.83          & 0.83 & 0.84 & 0.8  & \textbf{0.86} \\
\multicolumn{2}{r|}{Sent.Transformer Similarity} & \textbf{0.49} & 0.46 & 0.39 & 0.3           & 0.33 & 0.19 & 0.38 & 0.36          \\ \hline
\multicolumn{10}{c}{\cellcolor[HTML]{EFEFEF}Simulated Flowchart}                                                                    \\ \hline
\multicolumn{2}{r|}{BLEU}                        & 0.47          & 0.43 & 0.33 & \textbf{0.48} & 0.04 & 0    & 0.02 & 0.11          \\
\multicolumn{2}{r|}{ROUGE-L}                     & 0.54          & 0.52 & 0.52 & \textbf{0.56} & 0.18 & 0    & 0.12 & 0.4           \\
\multicolumn{2}{r|}{BERTScore-F1}                & 0.9           & 0.9  & 0.89 & \textbf{0.91} & 0.82 & 0.79 & 0.79 & 0.86          \\
\multicolumn{2}{r|}{Sent.Transformer Similarity} & 0.84          & 0.84 & 0.84 & \textbf{0.87} & 0.41 & 0.06 & 0.5  & 0.69              
\\ \bottomrule
\end{tabular}%
}
\caption{Experiment results for Flowchart Description task across the same 100 samples. The best-performing model is highlighted in \textbf{bolded}.}
\label{tab:exp100-sim}
\end{table*}

\begin{table*}[]
\resizebox{\textwidth}{!}{%
\begin{tabular}{rr|rrrrrrrrr}
\toprule\\
\multicolumn{2}{l|}{Evaluation Metrics} &
  Claude &
  GPT4V &
  Step-1V &
  \begin{tabular}[c]{@{}r@{}}Gemini\\ ProVision\end{tabular} &
  \begin{tabular}[c]{@{}r@{}}Gemini\\ ProVision(CoT)\end{tabular} &
  \begin{tabular}[c]{@{}r@{}}InternLM\\ -XComposer2-VL\end{tabular} &
  LLaVA16-34B &
  \begin{tabular}[c]{@{}r@{}}Qwen-VL\\ -chat\end{tabular} &
  \begin{tabular}[c]{@{}r@{}}DeepSeek\\ -VL-7B-chat\end{tabular} \\ \hline
\multirow{3}{*}{Link} & Precision & \textbf{0.35} & 0.23 & 0.14 & 0.21 & 0.23 & 0.01 & 0 & 0.01 & 0.02 \\
                      & Recall    & \textbf{0.26} & 0.22 & 0.15 & 0.2  & 0.22 & 0.01 & 0 & 0.01 & 0.02 \\
                      & F1        & \textbf{0.3}  & 0.22 & 0.14 & 0.21 & 0.22 & 0.01 & 0 & 0.01 & 0.02 \\ \hline
\multirow{3}{*}{Node} & Precision & \textbf{0.94} & 0.72 & 0.68 & 0.7  & 0.68 & 0.09 & 0 & 0.05 & 0.24 \\
                      & Recall    & \textbf{0.95} & 0.73 & 0.68 & 0.71 & 0.68 & 0.16 & 0 & 0.06 & 0.24 \\
                      & F1        & \textbf{0.94} & 0.72 & 0.68 & 0.7  & 0.68 & 0.11 & 0 & 0.05 & 0.24 
\\ \bottomrule
\end{tabular}%
}
\caption{Experiment results for Flowchart-to-Mermaid on Simulalted Flowcharts across the same 100 samples. The best-performing model is highlighted in \textbf{bolded}.}
\label{tab:exp100-mermaid}
\end{table*}

\subsection{Sample Model Responses for all tasks}
In this section, we showcase a selection of response examples from various models. These examples demonstrate how different models processed and responded to different Visual Question Answering (VQA) tasks across both scientific and simulated flowcharts within the FlowLearn dataset. The responses are displayed in Table \ref{tab:model_response1}, Table \ref{tab:model_response2}, and Table \ref{tab:model_response3}. For each VQA task, we aim to include at least one effective response and one less effective response to highlight the range of model capabilities and limitations.


\begin{table*}[]
\begin{tabular}{ll}
Dataset             & FlowLearn - Scientific Flowcharts                                                                                   \\
Example File        & 2204.00424v1-Figure6-1.png       \\ &
\begin{minipage}{.8 \textwidth}
      \includegraphics[width=0.8\textwidth]{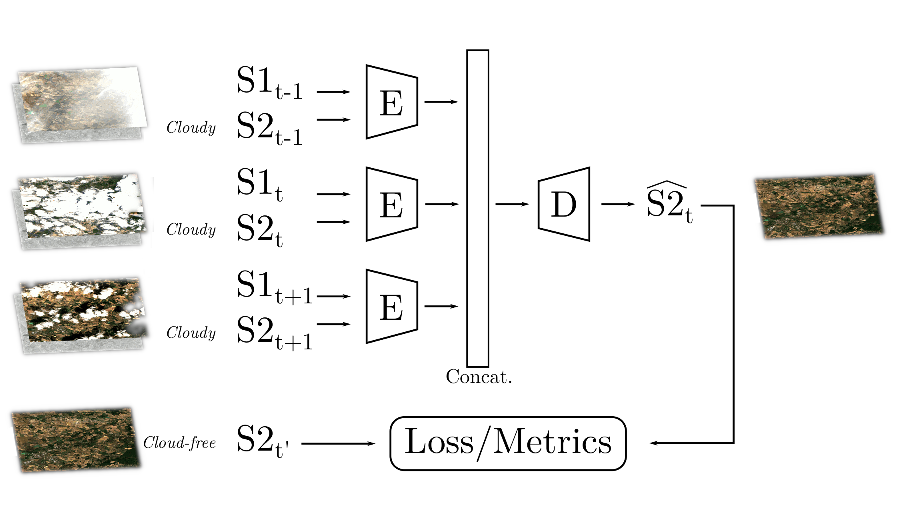}
    \end{minipage} 
\\
\rowcolor[HTML]{EFEFEF} 
Task                & Flowchart-to-Caption                                                                           \\
Reference &
  \begin{tabular}[c]{@{}l@{}}Figure 6. MSOPunet network. (S1t-1, S2t-1), (S1t, S2t) and (S1t+1, S2t+1) denotes the input pairs of optical and SAR \\ images. S2t' denotes the reference image used to compute the loss (during training) and the metrics (at inference time \\ from test data) from the reconstructed optical image S2t.\end{tabular} \\
  \hline
GPT4V &
  \begin{tabular}[c]{@{}l@{}}The answer is a flowchart depicting a process for handling satellite images to predict cloud-free images. The steps \\ involve the following:\\ \\ 1. Input of a pair of images at time t-1 (S1\_t-1 and S2\_t-1), labeled as "Cloudy."\\ 2. They are each passed through an encoder (E) to produce encoded representations.\\ 3. Another pair of images at time t with labels "Cloudy" for S1\_t and \color{red}{"Cloud-free" for S2\_t are also passed through} \\ \color{red}{the same encoder (E).}\\ 4. Another image at time t+1 (S2\_t+1) labeled as "Cloudy" is encoded similarly.\\ 5. All the encoded representations are concatenated.\\ 6. The encoded concatenation is passed through a decoder (D) to generate an output (S2\_t).\\ 7. Finally, this output is compared to a target image and assessed using Loss/Metrics to evaluate the performance \\ of the prediction.\end{tabular} \\
  \hline
InternLM-XComposer2-VL &
  \begin{tabular}[c]{@{}l@{}}The image is a flowchart that shows the architecture of 
  {\color{red}BERT} in pretraining. The model's task is to \color{red}{predict the masked}\\  {\color{red}{tokens}} from the representation vectors it learns at the positions of the input embeddings. The diagram also includes a \\ section on Loss/Metrics, which indicates how the loss is calculated and what metrics are used to evaluate the \\ performance of the model during training.\end{tabular} \\
\rowcolor[HTML]{EFEFEF} 
Task                & OCR                                                                                            \\
Reference           & Loss/Metrics                                                                                   \\\hline
GeminiProVision     & \color{teal}{Loss/Metrics}                                                                                   \\\hline
Qwen-VL-chat        & \color{red}{\begin{tabular}[c]{@{}l@{}}1\\ Picture 4: 2\\ The loss metrics are computed using\end{tabular}} \\\hline
LLaVA16-34B &
 \color{red}{ \begin{tabular}[c]{@{}l@{}}C1\end{tabular}} \\
\rowcolor[HTML]{EFEFEF} 
Task                & Statement: True                                                                                \\
Question Statement  & The output of 3 encoder layers are concatenated.                                               \\
Reference           & true                                                                                           \\
GPT4V               & \color{red}{false}                                                                                             \\
Claude              & \color{teal}{true}                                                                                            \\
\rowcolor[HTML]{EFEFEF} 
Task                & Statement: False                                                                               \\
Question Statement  & Cloudy image is used to calculate loss.                                                        \\
Reference           & false                                                                                          \\
GeminiProVision     & \color{teal}{false}                                                                                          \\
DeepSeek-VL-7B-chat & \color{red}{true}                                                                                          
\end{tabular}%
\caption{Models' responses to all VQA tasks on a Scientific Flowchart from the FlowLearn Dataset. Correct responses are highlighted in \textcolor{teal}{teal}, and incorrect responses are marked in \textcolor{red}{red}.}
\label{tab:model_response1}
\end{table*}

\begin{table*}[]
\begin{tabular}{ll}
Dataset &
  SimFlowchart-word \\
Example File &
  8394.jpeg \\
   &
\begin{minipage}{.7 \textwidth}
      \includegraphics[width=0.7\textwidth]{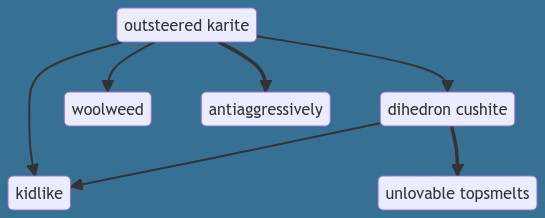}
    \end{minipage} 
\\
\rowcolor[HTML]{EFEFEF} 
Task &
  Flowchart-to-Mermaid \\
Reference &
  \begin{tabular}[c]{@{}l@{}}```mermaid\\ flowchart TB\\ entity0(outsteered karite)\\ entity1(dihedron cushite)\\ entity2(woolweed)\\ entity3(kidlike)\\ entity4(antiaggressively)\\ entity5(unlovable topsmelts)\\ entity0 --\textgreater entity1\\ entity0 --\textgreater entity2\\ entity0 --\textgreater entity3\\ entity0 ==\textgreater entity4\\ entity1 --\textgreater entity3\\ entity1 ==\textgreater entity5\\ ```\end{tabular} \\
  \hline
GeminiProVision &
 \color{teal}{\begin{tabular}[c]{@{}l@{}}```mermaid\\ flowchart LR\\ entity0(outsteered karite)\\ entity1(woolweed)\\ entity2(antiaggressively)\\ entity3(dihedron {\color{red}{cushtie}})\\ entity4(kidlike)\\ entity5(unlovable topsmelts)\\ entity0 --\textgreater entity1\\ entity0 --\textgreater entity2\\ entity0 --\textgreater \color{red}{entity3}\\ \color{red}{ entity1 --\textgreater entity4}\\ {\color{red}{entity3}} --\textgreater entity5\\ ```\end{tabular}} \\\hline
Qwen-VL-chat  (cropped)&
  \color{red}{\begin{tabular}[c]{@{}l@{}}1. outsteered karite 2. woolweed 3. antiaggresively 4. dihedron cushite 5. kidlike 6. unlovable topsmelts 7. karite \\ 8. woolweed 9. antiaggresively 10. dihedron cushite 11. kidlike 12. unlovable topsmelts 13. karite 14. woolweed \\ 15. antiaggresively 16. dihedron cushite 17. kidlike 18. unlovable topsmelts 19. karite 20. woolweed 21. antiaggresively \\ 22. dihedron cushite 23. kidlike 24. unlovable topsmelts 25. karite 26. woolweed 27. antiaggresively ...\end{tabular}} \\\hline
  \end{tabular}%
\caption{Models' responses to the first segment of VQA tasks on an evaluated Simulated Flowchart from the FlowLearn Dataset. Correct responses are highlighted in \textcolor{teal}{teal}, and incorrect responses are marked in \textcolor{red}{red}.}
\label{tab:model_response2}
\end{table*}

\begin{table*}[]
\begin{tabular}{ll}
Dataset &
  SimFlowchart-word \\
Example File &
  8394.jpeg \\
   &
\begin{minipage}{.7 \textwidth}
      \includegraphics[width=0.7\textwidth]{figures/8394.jpeg}
    \end{minipage} 
\\
\rowcolor[HTML]{EFEFEF} 
Task &
  Description \\
Reference &
  \begin{tabular}[c]{@{}l@{}}outsteered karite points to dihedron cushite. outsteered karite points to woolweed. outsteered karite points to \\ kidlike. outsteered karite points to antiaggressively. dihedron cushite points to kidlike. dihedron cushite points \\ to unlovable topsmelts.\end{tabular} \\ \hline
  LLaVA16-34B & \color{red}{\begin{tabular}[c]{@{}l@{}}The image contains a flowchart. Generate the description of the flowchart, reflecting the text nodes and arrows \\as depicted.\end{tabular}}\\ \hline
Claude &
  \color{teal}{\begin{tabular}[c]{@{}l@{}}The answer is:\\ \\ outsteered karite points to woolweed. \\ outsteered karite points to antiaggresively. \\ \color{red}{woolweed points to kidlike.}\\ \color{red}{antiaggresively points to dihedron cushite.}\\ dihedron cushite points to unlovable topsmelts.\end{tabular}} \\
\rowcolor[HTML]{EFEFEF} 
Task &
  OCR \\
Reference &
  outsteered karite \\
InternLM-XComposer2-VL &
 \color{red}{kidlike} \\
DeepSeek-VL-7B-chat &
  \color{red}{lansing} \\
Step-1V &
  \color{teal}{outsteered karite} \\
\rowcolor[HTML]{EFEFEF} 
Task &
  Num\_Nodes \\
Reference &
  6 \\
GPT4V &
  \color{teal}{6} \\
GeminiProVision &
  \color{red}{The total number of text nodes in the flowcharts is 13.} \\
\rowcolor[HTML]{EFEFEF} 
Task &
  Num\_Arrows \\
Reference &
  6 \\
Claude &
  \color{red}{7} \\
Step-1V &
  \color{red}{5} \\
\rowcolor[HTML]{EFEFEF} 
Task &
  AtoB Statement: True \\
Question Statement &
  Arrow points from node 'dihedron cushite' to node 'unlovable topsmelts'. \\
Reference &
  true \\
InternLM-XComposer2-VL &
  \color{teal}{true} \\
GPT4V &
  \color{red}{false} \\
\rowcolor[HTML]{EFEFEF} 
Task &
  AtoB Statement: False \\
Question Statement &
  Arrow points from node 'kidlike' to node 'outsteered karite'. \\
Reference &
  false \\
DeepSeek-VL-7B-chat &
  \color{teal}{false} \\
InternLM-XComposer2-VL &
  \color{red}{true} \\
\rowcolor[HTML]{EFEFEF} 
Task &
  Between AB Statement: False \\
Question Statement &
  Arrow exists between node 'dihedron cushite' and node 'kidlike'. \\
Reference &
  false \\
GPT4V &
  \color{teal}{false} \\
DeepSeek-VL-7B-chat &
  \color{red}{true} \\
\rowcolor[HTML]{EFEFEF} 
Task &
  Between AB Statement: True \\
Question Statement &
  Arrow exists between node 'dihedron cushite' and node 'woolweed'. \\
Reference &
  true \\
Claude &
  \color{red}{false} \\
Step-1V &
  \color{teal}{true}

\end{tabular}%
\caption{Models' responses to the second segment of VQA tasks on an evaluated Simulated Flowchart from the FlowLearn Dataset. Correct responses are highlighted in \textcolor{teal}{teal}, and incorrect responses are marked in \textcolor{red}{red}.}
\label{tab:model_response3}
\end{table*}

\subsection{GPT4V for Flowchart-to-Caption}
We conducted a sentence-level evaluation of GPT-4V's responses in the Flowchart-to-Caption task. We divided the 100 responses into 597 sentences. A PhD student specializing in NLP research was tasked with evaluating the responses. The table presents sample findings as discussed in the main text.





\begin{table*}[]
\begin{tabular}{l}

{\cellcolor[HTML]{F3F3F3}Related Finding: 1) Satisfactory responses 2)complex images 3) long-response}                                                  \\
{\cellcolor[HTML]{F3F3F3}Sample: 2101.11189v1-Figure2-1.png         }   \\
\begin{minipage}{.7 \textwidth}
      \includegraphics[width=0.8\textwidth]{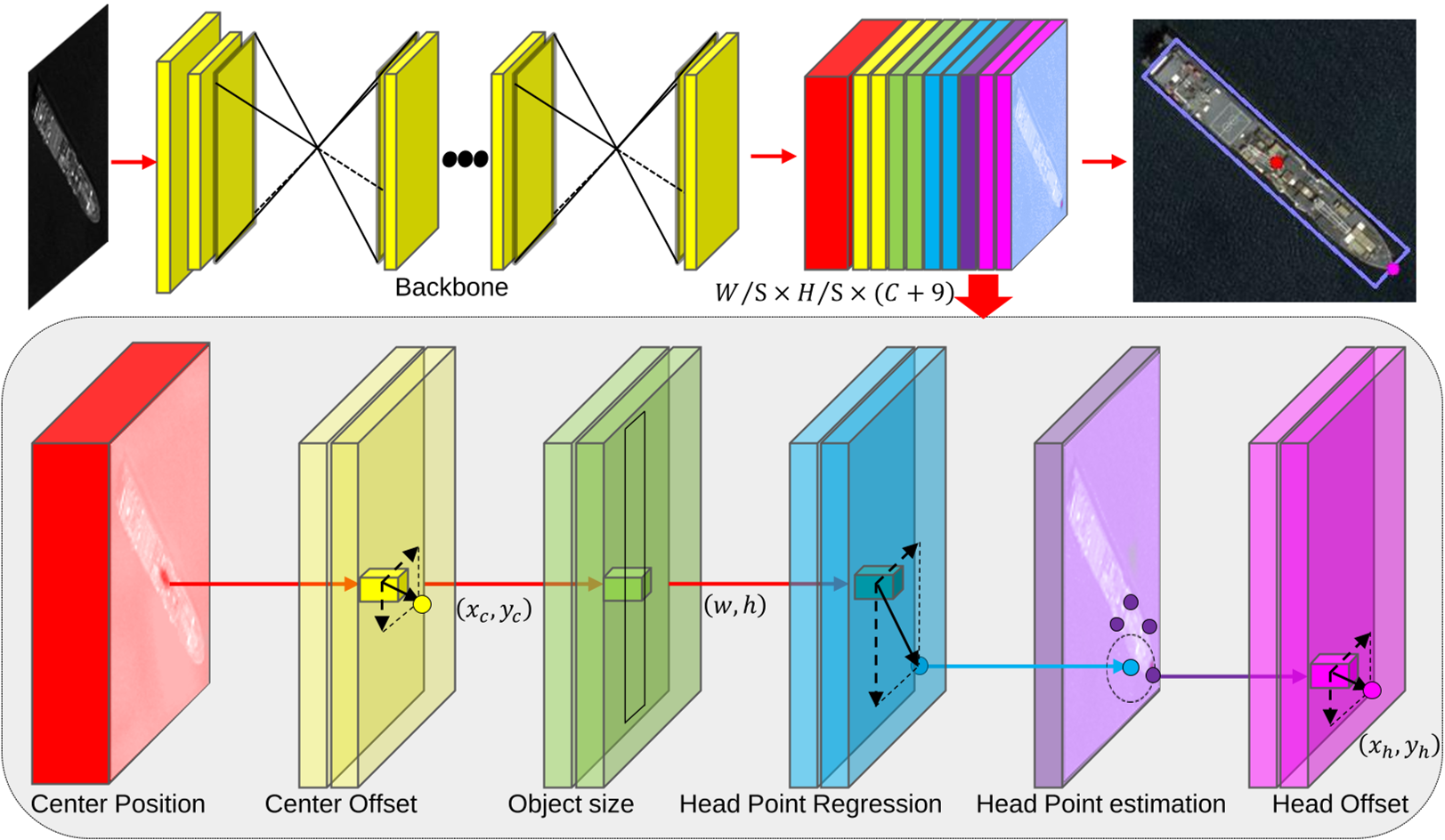}
    \end{minipage} 
\\
\textcolor{teal} {The image represents a deep learning architecture for object detection. }\\
\textcolor{teal} {It consists of the following sequence of key components:  }\\
\textcolor{teal} {1.Backbone:A neural network used for initial feature extraction from the input  }\\
\textcolor{teal} {2.A series of feature maps of different spatial resolutions, denoted by W/S × H/S × (C + 9). }\\
\textcolor{teal} {This likely represents feature maps with width and height scaled by a factor of S, containing C channels plus an additional 9 channels. }\\
\textcolor{teal} {3. These feature maps are processed through different branches for detecting various attributes of objects: }\\
\textcolor{teal} {- Center Position: Predicts the center (x\_c, y\_c) of an object.   }\\
\textcolor{teal} {- Center Offset: Compensates for errors in the predicted object center.   }\\
\textcolor{teal} {- Object size: Estimates the width (w) and height (h) of the object.  }\\
\textcolor{teal} {- Head Point Regression and Head Point Estimate: Localize a specific point on the object, usually the "head" point.       }\\
\textcolor{teal} {- Head Offset: Adjusts for errors in the head point prediction.    }\\
                                                                                                   \\
                                                                                                  \hline
                                                                                                  
{\cellcolor[HTML]{F3F3F3}Related Finding: 2)complex images 3) long-response 4) Logical but inaccurate descriptions, such as incorrectly interpreting the direction of arrows.}       \\
{\cellcolor[HTML]{F3F3F3}Sampling: 2204.01715v1-Figure7-1.png}  \\ 
\begin{minipage}{.8 \textwidth}
      \includegraphics[width=0.8\textwidth]{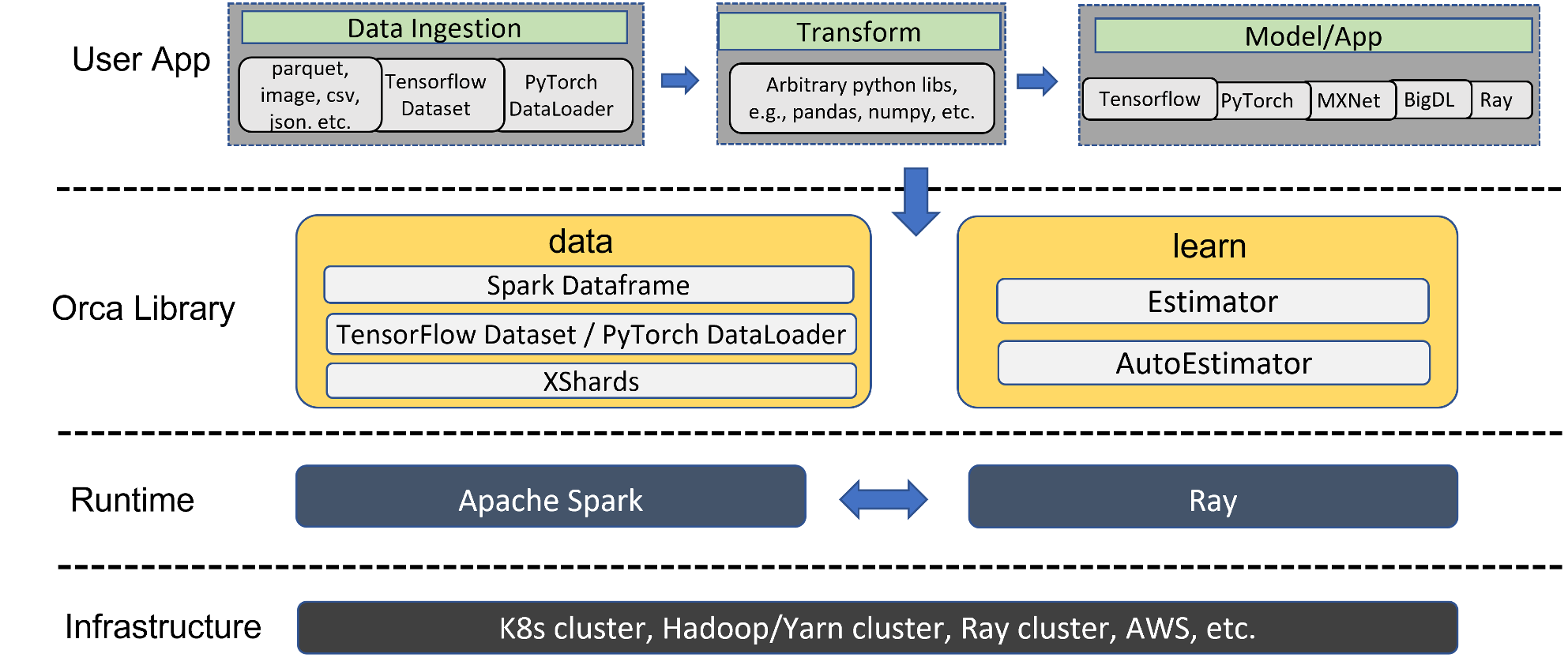}
    \end{minipage} 
\\
\textcolor{teal} {The answer is Figure 8: An overview of the system architecture. } \\
\textcolor{red} {The system is structured in three main layers: Infrastructure, Runtime, and User App.} \\
\textcolor{teal} {In the Infrastructure layer, there are options such as K8s cluster, Hadoop/Yarn cluster, Ray cluster, AWS, etc.    } \\
\textcolor{teal} {The Runtime layer consists of Apache Spark. } \\
\textcolor{teal} {In the User App layer, the flow starts from "Data Ingestion", moving to "Transform", and then to "Model/App".                            } \\
\textcolor{teal} {For "Data Ingestion" some mentioned inputs are image, csv, json, etc., and tools like Parquet, TensorFlow Dataset, or PyTorch Dataset can be used. } \\
\textcolor{red} {The "Transform" step includes tools such as Spark DataFrame, TensorFlow Dataset/PyTorch DataLoader, and XShards.    } \\
\textcolor{teal} {For the "Model/App", examples include PyTorch, BigDL, Ray, etc.    } \\
\textcolor{red} {\begin{tabular}[c]{@{}l@{}}The "data" component from the User App flows into the Orca Library in the Runtime layer, which then proceeds to the "learn" \\ component with Estimator and AutoEstimator.\end{tabular}} \\
\textcolor{red} {The information moves back and forth as indicated by the bidirectional arrows between "data" and "learn".               }                  \\
{\cellcolor[HTML]{F3F3F3}Sample: 2011.04105v1-Figure3.5-1.png}   \\                                             \begin{minipage}{.6 \textwidth}
      \includegraphics[width=0.6\textwidth]{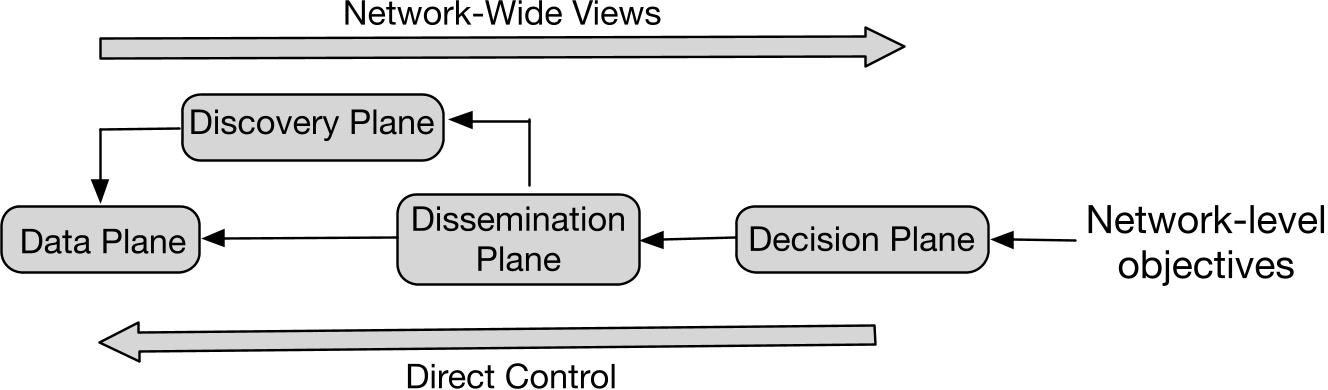}
    \end{minipage} 
\\                                                              \\
\textcolor{red} {The flowchart describes the structure of network views, with three primary planes and their interactions.     }                            \\
\textcolor{red} {\begin{tabular}[c]{@{}l@{}}Starting at the "Data Plane," there are two arrows branching out; one leads to the "Discovery Plane" and another flows directly to the \\ "Dissemination Plane."\end{tabular}} \\
\textcolor{red} {From the Discovery Plane, the flow moves to the Dissemination Plane.           }                                                           \\
\textcolor{red} {Then, from the Dissemination Plane, the process progresses to the "Decision }\\
\textcolor{teal} {The Decision Plane then aligns with "Network-level objectives. }                                                                           \\
\textcolor{red} {\begin{tabular}[c]{@{}l@{}}"There are also two broader categories depicted as horizontal arrows spanning the entire process: "Network-Wide Views" at the top \\ and "Direct Control" at the bottom, indicating overarching concepts that encompass the specific planes.\end{tabular} }\\
\end{tabular}%
\caption{GPT-4V's responses to the Flowchart-to-Caption tasks on Scientific Flowcharts from the FlowLearn Dataset, evaluated at the sentence level. Sentences are highlighted to indicate accuracy: correct sentences in \textcolor{teal}{teal} and incorrect sentences in \textcolor{red}{red}.}
\label{tab:my-table}
\end{table*}